\documentclass[sigconf, nonacm]{aamas}

\usepackage{balance} 
\usepackage{graphicx}

\setcopyright{ifaamas}
\acmConference[AAMAS '25]{Proc.\@ of the 24th International Conference
on Autonomous Agents and Multiagent Systems (AAMAS 2025)}{May 19 -- 23, 2025}
{Detroit, Michigan, USA}{A.~El~Fallah~Seghrouchni, Y.~Vorobeychik, S.~Das, A.~Nowe (eds.)}
\copyrightyear{2025}
\acmYear{2025}
\acmDOI{}
\acmPrice{}
\acmISBN{}

\acmSubmissionID{<<968>>}

\title{Certified Guidance for Planning with Deep Generative Models}

\author{Francesco Giacomarra}
\affiliation{
  \institution{University of Trieste}
  \city{Trieste}
  \country{Italy}}
\email{francesco.giacomarra@phd.units.it}

\author{Mehran Hosseini}
\affiliation{
  \institution{King's Colledge London}
  \city{London}
  \country{United Kingdom}}
\email{mehran.hosseini@kcl.ac.uk}

\author{Nicola Paoletti}
\affiliation{
  \institution{King's Colledge London}
  \city{London}
  \country{United Kingdom}}
\email{nicola.paoletti@kcl.ac.uk}

\author{Francesca Cairoli}
\affiliation{
  \institution{University of Trieste}
  \city{Trieste}
  \country{Italy}}
\email{francesca.cairoli@units.it}

\begin{abstract}

Deep generative models, such as generative adversarial networks (GANs) and score-based diffusion models,  have recently emerged as powerful tools for planning tasks and behavior synthesis in autonomous systems. Various guidance strategies have been introduced to steer the generative process toward outputs that are more likely to satisfy the planning objectives. These strategies avoid the need for model retraining but do not provide any guarantee that the generated outputs will satisfy the desired planning objectives. To address this limitation, we introduce \emph{certified guidance}, an approach that modifies a generative model -- without retraining it -- into a new model guaranteed to satisfy a given specification with probability $1$. We focus on Signal Temporal Logic (STL) specifications, which are rich enough to describe non-trivial planning tasks. Our approach leverages neural network verification techniques to systematically explore the generative models' latent spaces, identifying latent regions that are certifiably correct w.r.t. the STL property of interest. We evaluate the effectiveness of our method on four planning benchmarks using GANs and diffusion models. Our results confirm that certified guidance produces generative models that are always correct, unlike existing (non-certified) guidance methods.

\end{abstract}


\newcommand{\BibTeX}{\rm B\kern-.05em{\sc i\kern-.025em b}\kern-.08em\TeX}

\usepackage{graphicx} 
\usepackage{mathtools}
\usepackage{mathrsfs}
\usepackage{physics}
\usepackage{xcolor}
\usepackage{url}
\usepackage{amsmath,amsthm,amsfonts}
\usepackage[inline]{enumitem}  
\usepackage{booktabs}
\usepackage[capitalise]{cleveref}

\usepackage[toc,page]{appendix}

\setlist[itemize]{leftmargin=*}
\setlist[enumerate]{leftmargin=*}

\usepackage{array} 
\usepackage{algorithm}
\usepackage[noend]{algpseudocode}

\DeclareRobustCommand{\rchi}{{\mathpalette\irchi\relax}}
\newcommand{\irchi}[2]{\raisebox{\depth}{$#1\chi$}} 

\newcommand{\fc}[1]{{\color{green!50!black}[FC:#1]}}

\newcommand{\trainingset}{D_t}
\newcommand{\numtrain}{N_t}

\newcommand{\statespace}{S}
\newcommand{\statevar}{\mathbf{s}}
\newcommand{\sdim}{n}
\newcommand{\latentdim}{k}
\newcommand{\latentspace}{Z}
\newcommand{\latentvar}{\mathbf{z}}
\newcommand{\targetspace}{X}
\newcommand{\targetvar}{\mathbf{x}}
\newcommand{\horizon}{H}
\newcommand{\condhorizon}{h}
\newcommand{\condspace}{Y}
\newcommand{\condvar}{\mathbf{y}}

\newcommand{\property}{\phi}
\newcommand{\reward}{R}
\newcommand{\loss}{\mathcal{L}}
\newcommand{\ball}{\mathcal{B}}
\newcommand{\rew}{\mathcal{R}}

\newcommand{\nballs}{L}
\newcommand{\weights}{\theta}
\newcommand{\gen}{G_\weights}
\newcommand{\tgen}{\tilde{G}_\weights}
\newcommand{\robgen}{r^\property_\weights}
\newcommand{\boolgen}{b^\property_\weights}

\newtheorem{definition}{Definition}
\newtheorem{problem}{Problem}

\DeclareMathOperator*{\argmax}{arg\,max}

\Crefname{algorithm}{Alg.}{Algs.}
\Crefname{equation}{Eq.}{Eqs.}
\Crefname{figure}{Fig.}{Figs.}
\Crefname{table}{Tbl.}{Tbls.}
\Crefname{section}{\S\!}{\S\!}
\Crefname{subsection}{\S\!}{\S\!}
\Crefname{subsubection}{\S\!}{\S\!}
\Crefname{appendix}{\S\!}{\S\!}
\Crefname{theorem}{Thm.}{Thms.}
\Crefname{lemma}{Lem.}{Lems.}
\Crefname{proposition}{Prop.}{Props.}
\Crefname{problem}{Prob.}{Probs.}
\Crefname{remark}{Rem.}{Rems.}
\Crefname{definition}{Def.}{Defs.}
\Crefname{corollary}{Cor.}{Cors.}
\Crefname{example}{Ex.}{Exs.}
\Crefname{algorithm}{Alg.}{Algs.}

\begin{document}


\pagestyle{fancy}
\fancyhead{}


\maketitle 

\section{INTRODUCTION}\label{sec:intro}

Deep generative models (DMG) have become increasingly powerful and popular for tasks like data augmentation, planning, and behaviour synthesis. These models operate by transforming a standard distribution over a latent space into a more complex distribution over the target space, allowing for efficient sampling of high-dimensional data. 
Notable examples of state-of-the-art DGMs include Generative Adversarial Networks (GANs), Variational Autoencoders, and Score-based Diffusion Models. 

A key advantage of DGMs is their ability to generate samples that meet desired criteria, e.g., a planning task specification, without explicitly encoding these constraints or retraining the model. This is achieved by so-called \textit{guidance strategies}, which  
use a differentiable reward function to express the desired conditions and guide the generative process by optimizing the gradient of this reward. This technique parallels guided sampling, reinforcement learning, and planning algorithms, offering scalability, task composability, and effective non-greedy planning solutions~\cite{janner2022planning}.

Despite these advantages, guidance strategies are inherently limited by their inability to guarantee that the generated outputs will satisfy the desired constraints. When a requirement is unfeasible, the model may be pushed outside the data manifold, resulting in random or unreliable outputs that only coincidentally meet the specified conditions. This lack of robustness presents a significant challenge when using DGMs in safety-critical planning tasks.

To address these limitations, we introduce \textit{certified guidance}, a method that constrains the latent space of a DGM to ensure that generated samples always satisfy a given specification while preserving the relative likelihood of the original latent inputs. Certified guidance modifies the generative model, without retraining it, into a new model guaranteed to satisfy the specification with probability $1$. To do so, our method leverages neural network verification techniques to identify regions in the latent space of the DGM that are guaranteed to satisfy the specification. 

In particular, we consider specifications expressed in  Signal Temporal Logic (STL)~\cite{maler2004monitoring}, which provides a powerful formalism for defining complex temporal requirements, including planning tasks. In particular, our method uses STL both as a certification mechanism and as a differentiable reward function for guiding the generative process.  To our knowledge, certified guidance is the first method to combine neural network verification with deep generative models to certify temporal logic properties.

We demonstrate the effectiveness of certified guidance on four planning tasks, showing that our approach can certify both GAN and diffusion models and provide stronger correctness guarantees than existing (gradient-based) guidance methods, while providing samples representative of the original DGM distribution. 
\subsection*{Related Work}
\subsubsection*{Generative Models for Planning.}  Generative models, particularly those based on diffusion models, have gained popularity recently. For instance, \citet{Janner+22} introduced \emph{diffuser} for flexible behaviour synthesis using a denoising diffusion probabilistic model, whereby diffusers plan by iteratively refining randomly sampled noise. Decision diffusers~\cite{ajay2023conditional} 
are another popular approach where guidance is implemented by conditioning instead of latent space search. \citet{Dong+24} put forward DiffuserLite, which employs a planning refinement process to generate fine-grained trajectories by reducing the modelling of redundant information and increasing the decision-making frequency.

\subsubsection*{Verification of Probabilistic Models.} Closely related is the literature on single-step verification of probabilistic models. For instance, \cite{WickerLPK20, BattenHL24} on verification of Bayesian neural network. Specifically, we will use the pure and gradient-guided iterative expansion approach from \cite{BattenHL24} to search the latent space of the generative model.

\subsubsection*{Safety of Planners.} Most of the literature in this line of research focuses on the verification of sequential decision-making, where the decisions are usually made iteratively \cite{OrtheyCK24} using feedforward \cite{Akintunde+22}, recurrent \cite{HosseiniL23-RNN_Verif}, or Bayesian neural networks \cite{Wicker+24-BNN_ReachAvoid}. However, the safety verification of planners, especially those based on generative models, is an emerging line of research. Notably \citet{Xiao+23} introduced SafeDiffuser, which ensures diffusion probabilistic models satisfy specifications using a class of control barrier functions. Other work similar work include \cite{Sun+23,CairoliPB23-CP_STL,Kuipers+24-CP_MAS,LindemannCSP23-CP_Planning}, which use conformal prediction for uncertainty-aware planning. Nonetheless, to the best of our knowledge, the work presented here is the first to specifically study the safety verification of generative models for planning. closest to this work is \cite{Katz+22-Image_Planners}, which considers the verification of image-based neural network-based controllers, and to achieve this, it uses generative models for reducing the input dimension of such controllers. Nonetheless, this work does not consider the problem of verifying generative planners itself.

\section{BACKGROUND}\label{sec:background}

This section presents the key concepts needed to better understand the method detailed in \cref{sec:problem,sec:method}.

\subsection{Signal Temporal Logic (STL)}\label{subsec:stl}

Signal temporal logic (STL)~\cite{maler2004monitoring} is a formal language for specifying and monitoring the behaviour of physical
systems, including temporal constraints between events. 
STL
allows reasoning about dense-time, real-valued signals $\vec{\statevar}:\mathbb{T} \rightarrow S$, where $\mathbb{T}$ is the (continuous) time domain and $S\subseteq \mathbb{R}^n$ is the signal's space. 
It does so by mapping $\vec{\statevar}$ into a Boolean signal using atomic predicates of the form $\mu_g \equiv (g(\vec{\statevar}(t))>0)$ where $g: S \rightarrow \mathbb{R}$. 
STL properties are defined by the following syntax:
\begin{equation}\label{eq:stl_syntax}
    \varphi := \mathit{true}\ |\ \mu_g \ |\ \neg\varphi\ |\  \varphi\land\varphi\ |\ \varphi\ \mathcal{U}_{[a,b]}\varphi,
\end{equation}
where $[a,b]\subseteq \mathbb{T}$ is a bounded temporal interval. The \emph{until} operator $\varphi_1\ \mathcal{U}_{[a,b]}\varphi_2$ asserts that $\varphi_1$ must hold continuously until $\varphi_2$ becomes true at some point in the future.
From this essential syntax, we can define as usual other operators as follows: $false:=\neg true$, $\varphi \lor \psi := \neg(\neg\varphi\land\neg\psi)$, $\Diamond_{[a,b]} \varphi:= true\ \mathcal{U}_{[a,b]}\varphi$ and $\square_{[a,b]} \varphi:= \neg\Diamond_{[a,b]}\neg\varphi$, where $\Diamond_{[a,b]}\varphi$ and $\square_{[a,b]}\varphi$ denote respectively the \emph{eventually} and \emph{globally} operator. The first asserts that condition $\varphi$ will hold at some point in the time interval $[a,b]$, and the second that $\varphi$ holds at all times in the interval $[a,b]$.

\paragraph{Boolean semantics.}
The satisfaction of a formula $\varphi$ by a signal $\vec{\statevar}$ at time $t$ is defined by:
\begin{itemize}
   \item[-] $(\vec{\statevar},t) \models \mu_g \iff g(\vec{\statevar}(t))>0$; 
   \item[-] $(\vec{\statevar},t) \models \varphi_1 \land\varphi_2 \iff (\vec{\statevar},t) \models\varphi_1\land(\vec{\statevar},t) \models\varphi_2$;
   \item[-] $(\vec{\statevar},t) \models\neg\varphi \iff \neg((\vec{\statevar},t) \models\varphi))$;
   \item[-] $(\vec{\statevar},t) \models\varphi_1 \mathcal{U}_{[a,b]}\varphi_2 \iff \exists t'\in [t+a,t+b] \mbox{ s.t. }\\ (\vec{\statevar},t') \models\varphi_2\land \forall t''\in [t,t'), (\vec{\statevar},t'') \models\varphi_1$.

    \item[-] $(\vec{\statevar},t) \models \Diamond_{[a,b]}\varphi \iff \exists t'\in[t+a,t+b]  \mbox{ s.t. } (\vec{\statevar},t') \models\varphi$;
    \item[-] $(\vec{\statevar},t) \models \square_{[a,b]}\varphi \iff \forall t'\in[t+a,t+b]  (\vec{\statevar},t') \models\varphi$.
\end{itemize}
Given formula $\varphi$ and a signal $\vec{\statevar}$ over a bounded time interval, we can define the Boolean satisfaction signal as $\rchi^\varphi(\vec{\statevar},t)= 1$ if $(\vec{\statevar}, t)\models\varphi$ and $\rchi^\varphi(\vec{\statevar},t)= 0$ otherwise.
Monitoring the satisfaction of a formula is done recursively, by computing $\rchi^{\varphi_i}(\vec{\statevar},\cdot)$ for each sub-formula $\varphi_i$ of $\varphi$. The recursion is performed by leveraging the tree structure of the STL formula, where each node represents a sub-formula, in an incremental fashion so that the leaves are the atomic
propositions and the root represents the whole formula. Thus, the procedure goes bottom-up from atomic predicated to the top formula.

\paragraph{Quantitative semantics.} 
A useful feature of STL is that it admits a quantitative notion of satisfaction, called (space) \textit{robustness},  which quantifies how much a signal can be perturbed with additive noise before changing the truth value of a given property $\varphi$~\cite{donze2010robust}. It is defined as a function $R_\varphi$ such that:
\begin{align*}
   &R_{\mu_g} (\vec{\statevar},t) = g(\vec{\statevar}(t));\\ 
   &R_{\neg\varphi}(\vec{\statevar},t) = -R_{\varphi}(\vec{\statevar},t);\\
   &R_{\varphi_1\land\varphi_2}(\vec{\statevar},t) = \min (R_{\varphi_1}(\vec{\statevar},t),R_{\varphi_2}(\vec{\statevar},t));\\
  &R_{\varphi_1 \mathcal{U}_{[a,b]}\varphi_2}(\vec{\statevar},t) =
   \underset{t'\in [t+a,t+b]}{\sup}\left(\min\left(R_{\varphi_2}(\vec{\statevar},t'), \underset{t''\in [t,t']}{\inf}R_{\varphi_1}(\vec{\statevar},t'')\right)\right). 
\end{align*}
The sign of $R_\varphi$ indicates the satisfaction value. In particular, STL robustness is sound in the following sense~\cite{donze2010robust}:
 $R_\varphi(\vec{\statevar},t)>0\Rightarrow (\vec{\statevar},t)\models\varphi$ and $R_\varphi(\vec{\statevar},t)<0\Rightarrow (\vec{\statevar},t)\not\models\varphi$. 
As for the Boolean semantics, it is possible to generate monitors for the quantitative semantics automatically. 
The algorithm follows a similar bottom-up approach over the syntax tree of the formula. 

\subsection{Deep Generative Models}\label{subsec:dgm}



Every dataset can be considered a set of observations $\targetvar$ drawn from an unknown distribution $p(\targetvar)$. Generative models aim to learn a model that mimics this unknown distribution as closely as possible, i.e., learn a parametric distribution $p_\weights(\targetvar)$ that approximates $p(\targetvar)$ and allows for efficient sampling. 

A generative model acts as a distribution transformer, i.e., a map $\gen:\latentspace\to \targetspace$ transforming a simple distribution $p(\latentvar)$ over a latent space $\latentspace$ into a complex distribution $p_\weights(\targetvar)\approx p(\targetvar)$ over the target space $\targetspace$. In other words, they generate data from noise. 
Training the generative model requires minimising a differentiable loss function $\loss$, 
which measures the difference between true and generated samples for each condition $\condvar$ in $\trainingset$. 
The model architecture, its loss and thus its training procedure depend on the generative model selected. In this paper, we focus on Generative Adversarial Networks (GAN) and Denoising Diffusion models (DIFF). Details about such models are provided in \cref{app:dgm}. 

For planning applications, we are interested in generating stochastic trajectories conditioned, for instance, on the starting position of the system. 
To approximate such conditional distribution $p(\targetvar\mid \condvar)$, we use 
\textit{conditional deep generative models (CDGM)}, which can be seen as maps $\gen: \latentspace\times \condspace\to \targetspace$, where $\condspace$ is the conditioning space. The main difference lies in defining a conditional DGM  $\gen:\latentspace\times \condspace\to \targetspace$ that transforms a simple conditional distribution $p(\latentvar\mid \condvar)$ over the latent space $\latentspace$ into a complex distribution $p_\weights(\targetvar\mid \condvar)$ over the target space $\targetspace$ that approximates $p(\targetvar\mid \condvar)$.
Both GAN and DIFF support such a {conditional} formulation. 

\subsection{Guided sampling from DGM}\label{subsec:guidance}
 
As described before, the generative process of DGM is typically controlled through conditioning. Conditioning requires that the model is built from the ground up to accept a particular modality of conditions from the user, be it descriptive text, class labels, etc. While conditioning is a powerful tool, it results in models that are hand-cuffed to a single conditioning modality. If another modality is required, a new model needs to be trained, often from scratch. Unfortunately, the high cost of training makes this prohibitive for most users. A more flexible approach to controlling the generation process is \textit{guidance}~\cite{bansal2023universal}. 

Guidance enables generative models to be controlled by arbitrary modalities without retraining any components. The generative model $\gen$ is paired with a real-valued reward function $\rew: \targetspace\to\mathbb{R}$ measuring how much a criterion of interest is met by the generated sample. In our context, we could guide the model to generate trajectories that maximize the satisfaction of a planning specification. 

Let us define the real-valued function $r_\weights = \rew (\gen(\latentvar))$, which associates a reward to each generated sample. The guidance algorithm aims to search for the optimal latent input that satisfies:
\begin{equation}\label{eq:opt_search_bg}
    \latentvar^* = \argmax_\latentvar r_\weights (\latentvar).
\end{equation}
If $\rew$ is differentiable, one could compute $\nabla_\latentvar r_\weights(\latentvar)$, the gradient of $r_\weights$ w.r.t. $\latentvar$ via automatic differentiation and then leverage gradient-based optimization techniques to approximately solve~\eqref{eq:opt_search_bg}, i.e. to search for the latent input $\latentvar^*$ that generate the sample $\targetvar^* = \gen (\latentvar)$ associated with the highest reward. 

%
\subsection{Neural Network Verification}
Several methods for verifying the robustness of neural networks against input perturbations have been put forward in recent years. The goal here is to compute
\begin{equation*}
  f_L := \min_{\targetvar\in \ball (\targetvar_0,\varepsilon)} f(\targetvar)
  \quad \mbox{and} \quad
  f_U := \max_{\targetvar\in \ball (\targetvar_0,\varepsilon)} f(\targetvar).
\end{equation*}
which are lower and upper bounds for the output of a neural net \(f\) on an \(\varepsilon\)-ball (w.r.t. an \(\ell_p\)-norm) \(\ball (\targetvar_0,\varepsilon)\) around a given input \(\targetvar_0\).

Broadly speaking, approaches for solving this problem can be divided into \emph{complete} and \emph{incomplete} methods. Complete methods compute the exact values of \(f_L\) and \(f_U\) while incomplete methods compute a lower bound \(\tilde{f}_L \leq f_L\) and an upper bound \(\tilde{f}_U \geq f_U\).

A key difference between complete and incomplete methods is that incomplete methods use relaxation to approximate non-linearities in \(f\). In contrast, complete methods aim to compute exact upper and lower bounds for all neurons in the network. As such, complete approaches are usually more costly and less scalable than incomplete approaches. On the other hand, even though incomplete approaches tend to be more scalable, for larger models they result in extremely loose bounds that are not usable. Some state-of-the-art tools use a mixture of techniques from complete and incomplete methods to offer even more scalable approaches.

In our experiments, we use the Auto-LiRPA verifier~\cite{xu2020automatic}, see \cref{subsubsec:Implementation} for details. Nonetheless, the framework presented here is independent of the underlying neural network verification tool as long as it can derive sound bounds. 



\section{PROBLEM FORMULATION}\label{sec:problem}
Our focus is to certify that the realizations of a deep generative model (DGM) satisfy some planning goal expressed as a signal temporal logic (STL) requirement. More precisely, given a DGM $(\gen, p(\latentvar))$ -- uniquely identified by the latent distribution $p(\latentvar)$ and the pre-trained generator $\gen$ -- we define the \emph{probabilistic satisfaction} of an STL property $\property$ as the probability that for {$\latentvar\sim p_\property(\latentvar)$} the output of the DGM, $\gen(\latentvar)$, satisfies $\property$, i.e., $\rchi_\property (\gen(\latentvar))  = 1$, where $\rchi_\property$ denote the Boolean STL semantics.
\begin{definition}[Satisfaction Probability]\label{def:prob_sat}
Let $(\gen, p(\latentvar))$ be a DGM and $\property$ an STL property. 
The \emph{satisfaction probability} of $\property$ is defined as 
\begin{equation}
    P((\gen, p(\latentvar))\models \property) := Pr_{\latentvar\sim {p}(\latentvar)}\left(
\rchi_\property(\gen(\latentvar)) = 1
    \right).
\end{equation}
\end{definition}
 $\gen$ being a deterministic map, the stochasticity of the DGM is uniquely determined by the latent distribution {$p(\latentvar)$}. The goal of certified guidance is to derive from $p(\latentvar)$ a new 
property-specific latent distribution $p_\property(\latentvar)$ such that every realization of the DGM $(\gen, p_\property(\latentvar))$ satisfies $\property$.

\begin{problem}[Certified Guidance]\label{prbl:cert_guidance}
    For a DGM $(\gen,p(\latentvar))$, STL property $\property$, the \emph{certified guidance} problem is finding a new latent distribution $p_\property(\phi)$ for the DGM such that
    \begin{enumerate}
        \item $\property$ is always satisfied, i.e., 
        \begin{equation}\label{eq:certification}
            P\big((\gen,p_\phi(\latentvar)) \models \phi\big)= 1,
        \end{equation}
        
        \item the relative likelihood of the latent points is preserved, i.e., 
        \begin{equation}\label{eq:ratio}
            \forall \latentvar_1, \latentvar_2 \in \mathrm{supp}(p_\property), \ \frac{p_\property(\latentvar_1)}{p_\property(\latentvar_2)}=\frac{p(\latentvar_1)}{p(\latentvar_2)}.
        \end{equation}
    \end{enumerate}
\end{problem}
The first requirement, arguably the most important, states that the new DGM $(\gen,p_\phi(\latentvar))$ is certified, i.e., it satisfies property $\property$ with probability $1$. The second requirement states that any two latent points under $p_\property$ maintain the same relative importance between them as under $p$. This condition ensures that the original distribution, and so, the data distribution, is captured by $p_\property$. 
Note that \ref{prbl:cert_guidance} is formulated for a DGM $(\gen,p(\latentvar))$, but it can be naturally extended to conditional DGMs $(\gen,p(\latentvar\mid \condvar))$ for fixed values of the conditioning variable $\condvar$. 

A natural solution is to define $p_\property$ by restricting $p$ to a region of the latent space that provably satisfies $\property$. 
Following a line of reasoning similar to~\cite{WickerLPK20}, we now introduce (maximal) $\property$-satisfying sets as a tool to compute the satisfaction probability of $\phi$. 

\begin{definition}[Maximal $\property$-satisfying set]\label{def:sat_set}
   Given a DGM $(\gen,p(\latentvar))$, the maximal set of latent inputs for which the corresponding generated
trajectory satisfies property $\property$, namely the maximal $\property$-satisfying set, is defined as $\overline{B} =\{\latentvar\in \mathbb{R}^\latentdim|\rchi_\property (\gen (\latentvar)) = 1\}$. Furthermore, we say that $B$ is a $\property$-satisfying set of latent inputs iff $B\subseteq \overline{B}$.  
\end{definition}

The following proposition is a consequence of Definition~\ref{def:prob_sat} and~\ref{def:sat_set}. 
\begin{proposition}\label{prop:prob}
Let $\overline{B}$ be the maximal $\property$-satisfying set and $(\gen, p(\latentvar))$ be corresponding DGM. Then, it holds that
    \begin{equation}
    P\big((\gen,p(\latentvar)) \models \phi\big) = \int_{\overline{B}} p(\latentvar)d\latentvar.
\end{equation}
\end{proposition}

\cref{prop:prob} simply translates the computation of the satisfaction probability from the function space to an integral computation on the latent space. Computing $P((\gen,p(\latentvar)) \models \phi)$ reduces to computing the maximal set $\overline{B}$ of latent inputs for which the corresponding generated trajectory satisfies property $\property$. However, we seek to find a latent distribution $p_\property(\latentvar)$ such that $P((\gen,p_\property(\latentvar)) \models \phi) = 1$.

For a set of latent points $B$, let us use the following notation 
\begin{equation}\label{eq:p_given_B}
p(\latentvar|\latentvar\in  B) = \dfrac{p(\latentvar)\cdot \mathbf{1}(\latentvar \in B)}{p(B)}
\end{equation}
to denote the distribution obtained by restricting the support of $p$ to $B$, where $\mathbf{1}(\cdot)$ is the indicator function.  
As a consequence of Proposition~\ref{prop:prob}, if we define
$
p_\property(\latentvar) := p(\latentvar|\latentvar\in  \overline{B})$
it follows that 
$$P\big((\gen,p_\property(\latentvar)) \models \phi\big) = \int_{\overline{B}} p(\latentvar|\latentvar\in  \overline{B})d\latentvar = 1.$$

Identifying the maximal $\property$-satisfying set $\overline{B}$ is a non-trivial and generally infeasible task. Typically, we can identify a smaller $\property$-satisfying set  $B \subseteq \overline{B}$. If we define
$
p_\property(\latentvar) := p(\latentvar|\latentvar\in  B) $
it follows, as before, that 
\begin{equation}
    P\big((\gen,p_\property(\latentvar)) \models \phi\big) = \int_{B} p(\latentvar|\latentvar\in  B)  )d\latentvar = 1.
\end{equation}


\begin{corollary}\label{cor:satprob}
    Let $B$ be a $\property$-satisfying set. Then, it holds that 
    \begin{equation}\label{eq:corrolary}
P\big((\gen,p(\latentvar | \latentvar\in B)) \models \phi\big) = 1.
    \end{equation} 
    Moreover, by~\eqref{eq:p_given_B}, for all $\latentvar_1, \latentvar_2 \in B$, we have that
    $$
    \frac{p(\latentvar_1\mid \latentvar_1\in B)}{p(\latentvar_2\mid \latentvar_2\in B)} = \frac{p(\latentvar_1)}{p(\latentvar_2)}.
    $$ 
\end{corollary}



\cref{cor:satprob} guarantees that certified guidance is equivalent to finding a certified $\property$-satisfying set 
$B\subseteq \overline{B}$ and restricting the support of the original latent distribution $p(\latentvar)$ to $B$ resulting in a new  latent distribution $p_\property(\latentvar) = p(\latentvar | \latentvar \in B)$. 

We construct $B$ as the union of $M$ disjoint hyper-rectangles $B = \bigcup_{i=1}^M B_i$, where each $B_i = [l_1^i,u_1^i]\times\cdots\times [l_\latentdim^i,u_\latentdim^i]$ is a $\property$-satisfying set. This choice has two main advantages. First, a hyper-rectangle can be expressed as an $L^{\infty}$-ball, which can be certified by existing neural network verification tools. Second, the resulting distribution $p_{\property}(\latentvar)=p(\latentvar\mid \latentvar \in B)$ has an analytical form which makes sampling efficient. In contrast, implementing $p_{\property}$ by rejection sampling would be highly inefficient if $B$ has an overall low probability. Recall that the original latent distribution $p_{\property}(\latentvar)$ is a multi-variate isotropic standard normal, i.e., $\mathcal{N}(0,\mathbf{I}; \latentvar)$, meaning that its density can be factored as $p_{\property}(\latentvar)= \prod_{j=1}^\latentdim \mathcal{N}(0,1; z_j)$ where $z_j$ is the $j$-th component of $\latentvar$. 
This allows us to express the density $p_{\property}(\latentvar)$ as a mixture of truncated normal distributions, as follows:
\begin{equation}\label{eq:truncated_normal}
    p_{\property}(\latentvar) = \sum_{i=1}^M \dfrac{p(B_i)}{p(B)} \prod_{j=1}^\latentdim \mathcal{N}(0,1,l_j^i,u_j^i; \latentvar_j),
\end{equation}
where $\mathcal{N}(0,1,a,b; x)$ is the density of the standard normal at $x$ truncated within the interval $[a,b]$, and 
\begin{equation}\label{eq:p_Bi}
p(B_i) = \prod_{j=1}^\latentdim \frac{1}{2}\left(
\mathit{erf}\left(-\frac{l^i_j}{\sqrt{2}}\right)-\mathit{erf}\left(-\frac{u^i_j}{\sqrt{2}}\right)
\right)
\end{equation}
is the probability of the set $B_i$ w.r.t. the original latent $p$ (derived from the CDF of the standard normal), with $p(B)=p\left(\bigcup_{i=1}^M B_i\right) = \sum_{i=1}^M p(B_i)$ being the overall probability of $B$ (because the sets $B_i$s are mutually disjoint).
Next, we present a method to generate $\property$-satisfying sets $B$ of latent inputs. 

\subsection{Checking $\property$-Satisfying Sets}

We now describe how to check whether a given set, 
$B$, in the latent space, is such that $B\subseteq\overline{B}$, that is, we want to check whether $\rchi_\property (\gen(\latentvar)) = 1$, $\forall \latentvar\in B$.
This is equivalent to checking:
\begin{equation}\label{eq:min_lb}
    \min_{\latentvar\in B} \rchi_\property (\gen(\latentvar)) = 1.
\end{equation}

The verifier can be used to find a lower bound on the solution of the problem posed by Eq.~\eqref{eq:min_lb}.
Given a bounding box $B$ in the latent space of a DGM, the NN verifier propagates $B$ through the deterministic function $b^\property_\weights := \gen\circ \rchi_\property$ to obtain bounds over the Boolean satisfaction values, i.e. the set $\{0,1\}$. If both the lower and the upper bound of the output box are equal to $1$, then $B$ is a $\property$-satisfying set of latent inputs. 

In \cref{sec:method}, we describe a strategy to efficiently search for $\property$-satisfying sets, which employs gradient-based search to identify candidate $B$'s and iteratively expands them as long as they satisfy $\property$. 

\section{CERTIFIED GUIDANCE FOR PLANNING}\label{sec:method}

Before discussing in detail our certified guidance algorithm, we briefly introduce how to represent a planning problem with DGMs. 
DGMs are used to predict the trajectories (state sequences) of a stochastic system evolving over discrete (or discretized) time. The state at a specific instant of time can be described as a $\sdim$-dimensional vector $\statevar\in\statespace\subseteq\mathbb{R}^\sdim$. In particular, 
we use CDGMs (introduced in \cref{subsec:dgm}) to approximate the distribution $p(\targetvar\mid\condvar)$ of future state sequences $\targetvar \in S^{\horizon}$ given a conditioning prefix $\condvar\in S^{\condhorizon}$, for some sequence lengths $\horizon$ and $\condhorizon$. To approximate $p(\targetvar\mid\condvar)$, CDGMs use realizations of the stochastic systems (obtained via simulation or through direct measurements) as their training dataset: $\trainingset = \{(\condvar_i,\targetvar_i)\mid i= 1,\dots \numtrain\}.$
We denote with $\vec{\statevar}\in S^{\condhorizon+\horizon}$ the trajectory obtained by concatenating the prefix and target sequences, with $\vec{\statevar}(t)\in S$ being the state of $\vec{\statevar}$ at time $t \in \{0,\ldots,\condhorizon+\horizon-1\}$.

We now illustrate our solution method for the certified guidance problem, stated in \cref{prbl:cert_guidance}. Our solution produces a new latent distribution $p_\property(\latentvar|\condvar)$ of the generative model, such that the resulting CDGM $(\gen, p_\property(\latentvar|\condvar))$ offers verified satisfaction of $\property$ for all its realizations. We need to define a strategy to build the 
certified $\property$-satisfying sets 
$B_1,\dots, B_M$ to restrict the support of $p(\latentvar|\condvar)$.
The starting point is to consider a fixed condition ${\condvar}$,  pick a latent input $\latentvar_*$ and construct a bounding box $\mathcal{B}_p(\latentvar_*,\varepsilon)$ around $\latentvar_*$, where $\mathcal{B}_p(\latentvar_*,\varepsilon)$ denotes an $\varepsilon$-perturbation around $\latentvar_*$ with norm $L^p$. These bounding box in the latent space will be then certified by the neural network verification algorithm.  

We leverage reward-guided sampling (introduced in \cref{subsec:guidance}) to search for good candidates, or pivot points, for $\latentvar_*$. 
To this purpose, we use the quantitative STL semantics $R_\property$ (introduced in \cref{subsec:stl}) as a differentiable measure of reward $\rew$ to guide the DGM sampling. Let $\reward_\property(\vec{\statevar})$ denote the reward associated to the target trajectory $\vec{\statevar}$ w.r.t. $\property$. High positive rewards indicate satisfaction of the property $\property$, whereas low negative rewards indicate violation. 
Given a condition $\condvar$ and a pre-trained generative model $\gen$, we can define 
$$
\robgen(\latentvar, \condvar) := \reward_\property(\tgen(\latentvar, \condvar)),
$$
 which, for a fixed $\condvar$, is a function of $\latentvar$ alone. We stretch the notation a little and define $\tgen(\latentvar, \condvar)$ as the model which outputs trajectories in $\statespace^{\horizon+h}$ resulting from the concatenation of the condition $\condvar\in \statespace^h $ and the generated target $\gen(\latentvar, \condvar) \in \statespace^\horizon$.  
Searching for a pivot point can be seen as searching for the latent value $\latentvar^*$ whose corresponding generated trajectory maximizes the satisfaction of $\property$. That is, we solve the following optimization problem:
\begin{equation}\label{eq:optimization}
    \latentvar^* = \argmax_\latentvar
    \robgen(\latentvar,\condvar),
\end{equation}
which can be solved with gradient ascent techniques.
In practice, we randomly choose $\nballs$ latent points, $\latentvar_0^1,\dots,\latentvar_0^\nballs$ and run gradient ascent long enough to find $\latentvar_*^1,\dots,\latentvar_*^\nballs$. 
We choose a small positive $\varepsilon$ and build $\mathcal{B}_p(\latentvar_*^i,\varepsilon)$. 
Then, for each $i$, we iteratively increase $\varepsilon$ until $\mathcal{B}_p(\latentvar_*^i,\varepsilon)$ remains a $\phi$-satisfying set. 

\begin{figure}
    \centering
    \includegraphics[width=0.75\linewidth]{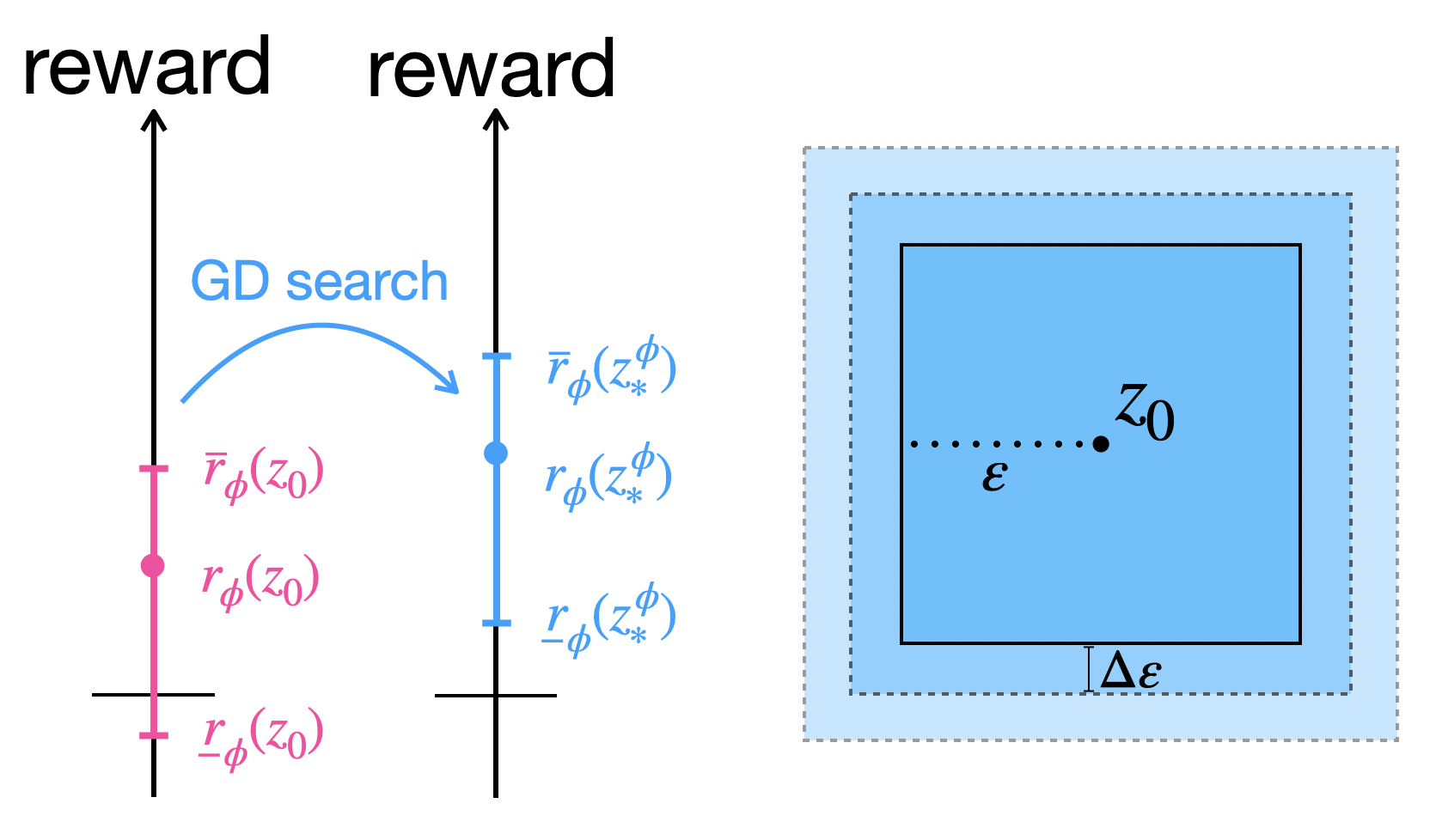}
    \caption{Certified Robust Generation: gradient ascent strategy (left) and iterative expansion of perturbation balls (right).\vspace{-.5cm}}
    \label{fig:certification}
\end{figure}

To determine whether $\mathcal{B}_p(\latentvar_*,\varepsilon)$ is a $\property$-satisfying set, we need to find a lower bound $\underline{\boolgen} (\latentvar_*, {\condvar})$
for 
$\boolgen (\latentvar_*, {\condvar}) = \rchi_\property(\tgen((\latentvar_*, {\condvar}))$, where $\rchi_\property$ denotes the Boolean semantics, i.e., $\rchi_\property (\vec{\statevar}) = 1$ 
if $\vec{\statevar}$ satisfies $\property$ and $\rchi_\property (\vec{\statevar}) =0$ otherwise.
Formally, $\underline{\boolgen} (\latentvar_*, {\condvar})$ is such that
\begin{equation}\label{eq:lb_ub}
   \forall \latentvar\in \ball (\latentvar_*,\varepsilon)\qquad 
\underline{\boolgen} (\latentvar_*, {\condvar}) \le \boolgen (\latentvar, {\condvar}).
\end{equation}
Then, we check whether $\underline{\boolgen} (\latentvar_*, {\condvar}) =1$, which implies that $\property$ is always satisfied by all latent points in the $\varepsilon$-ball $\ball (\latentvar_*,\varepsilon)$. 
Otherwise, the region cannot be certified, and $\mathcal{B}_p(\latentvar_*,\varepsilon)$ is disregarded. 
Importantly, when $\mathcal{B}_p(\latentvar_*,\varepsilon)$ is instead certified, we can \emph{iteratively increment} $\varepsilon$ (\cref{fig:certification}-right) until the maximum perturbation we successfully certify. Indeed, we are interested in certifying the largest volume possible in the latent space.

\paragraph{Heterogenous increments of $\underline{\varepsilon}$} The perturbation $\underline{\varepsilon}$ can be incremented heterogeneously along the $\latentdim$ latent dimensions. A similar approach has been introduced in \cite{BattenHL24} to certify the probabilistic robustness of Bayesian neural nets. The rationale behind this approach is to expand the hyper-rectangle more in the directions that provide a significant contribution to the increase of the STL robustness, $\underline{\varepsilon}' = \underline{\varepsilon}+\alpha\left(\frac{\nabla_\latentvar \robgen (\latentvar)}{\| \nabla_\latentvar \robgen (\latentvar)\|}\right)$, where  $\alpha$ is some ad-hoc constant.

\subsection{Algorithm}\label{sec:algorithm}


Algorithm~\ref{alg:prob_sat} shows our approach to building a certified DGM. For simplicity of notation, we illustrate the algorithm in the non-conditional case, i.e., using $p(\latentvar)$ instead of $p(\latentvar \mid \condvar)$. In lines 1–13, the algorithm computes $B$, which is the union of $\property$-satisfying sets of latent inputs
We build $B$ as a union of hyper-rectangles (line 13), each of which is $\property$-satisfying. Each hyper-rectangle is generated as follows: we sample a random latent input $\latentvar_0^i$ (line 6) from which we start a GD-search of a local optimum $\latentvar_*^i$ (line 7). This process is repeated $\nballs$ times to generate multiple $\property$-satisfying hyper-rectangles in disconnected regions of the latent space. In lines 8-11, we check if the given hyper-rectangle is $\property$-satisfying by using NN verification, and we also check that the region does not overlap with others (lines 9 and 12). We iteratively increment $\varepsilon$ as long as the $\property$-satisfying and the non-overlapping properties are met (lines 9-11). All the $M \leq \nballs $ $\property$-satisfying non-overlapping hyper-rectangles are added to $B$ (lines 12-13).
We then use this set of hyper-rectangles to build our certified latent distribution $p_\property(\latentvar)=p(\latentvar|\latentvar \in B)$, which is defined as a truncation of the original latent distribution $p(\latentvar|\condvar)$ over the $M$ disjoint hyper-rectangles, as explained in Eq.~\ref{eq:truncated_normal}. To do so, our algorithm computes the probability $p(B_i)$ of each box and the 
overall probability of $B$, $p(B)= \sum_{i=1}^M p(B_i)$, in lines 14-16. For numerical stability, these probabilities are computed on a logarithmic scale.



\begin{algorithm}
\caption{Certified Guidance for DGMs}\label{alg:prob_sat}
\begin{algorithmic}[1]
\State \textit{Input:} DGM $(\gen, p(\latentvar))$, STL property $\property$, max number of regions $L$, perturbation hyperparameters $\varepsilon_0, \Delta\varepsilon, \alpha$.

\State \textit{Output:} Certified DGM $(\gen, p_\property(\latentvar) = p(\latentvar | \latentvar \in B))$.

\State $B \gets \emptyset$
\For{$i=1:L$}
    \State $\varepsilon \gets \varepsilon_0$
    \State $\latentvar_0^i \sim p( \latentvar)$
    \State $\latentvar_*^i \gets \textsc{GD}(-r^\property_\weights(\latentvar),\latentvar_0^i,\gamma)$
        \Comment{\texttt{Gradient-based search}}
    \State $\underline{b}_\property,\overline{b}_\property \gets \textsc{NNVerification}\left(r_\weights(\latentvar), \mathcal{B}(\latentvar_*^i,\varepsilon)\right)$ \Comment{\texttt{Lower and Upper bounds over Boolean Satisfaction $\rchi_\property$}}
    \While{$\underline{b}_\property = 1\ \land \mathcal{B}(\latentvar_*^i,\varepsilon)\cap B = \emptyset$}
        \State $\varepsilon \gets \varepsilon + \Delta\varepsilon$ \Comment{\texttt{Heterogeneous:} $\varepsilon \gets \varepsilon + \alpha\nabla_\latentvar r_\theta^\phi(\latentvar)$}
        \State $\underline{b}_\property,\overline{b}_\property \gets \textsc{NNVerification}\left(r_\weights(\latentvar), \mathcal{B}(\latentvar_*^i,\varepsilon)\right)$
    \EndWhile
    \If{$\underline{b}_\property = 1\land \mathcal{B}(\latentvar_*^i,\varepsilon)\cap B = \emptyset$}
    \State $B\gets B\cup   \mathcal{B}(\latentvar_*^i,\varepsilon)$
    \EndIf

\EndFor


\If{$B\ne\emptyset$}\Comment{$p_\property(\latentvar) := p(\latentvar\mid \latentvar\in B)$}
\For{$i\gets 1:M$} \Comment{\texttt{$M$: \# hyper-rectangles in $B$}}
\State $ p(B_i) \gets \prod_{j=1}^\latentdim \frac{1}{2}\left(
\mathit{erf}\left(-\frac{l^i_j}{\sqrt{2}}\right)-\mathit{erf}\left(-\frac{u^i_j}{\sqrt{2}}\right)\right)$
\Comment{Eq.~\ref{eq:p_Bi}}
\EndFor
$p(B) \gets \sum_{i=1}^M p(B_i)$

\EndIf

\end{algorithmic}
\end{algorithm}

\section{EXPERIMENTAL RESULTS}\label{sec:experiments}

We test the proposed certification procedure across a variety of planning problems, encompassing both 2D and 3D environments. Our experimental framework employs deep generative models, specifically Wasserstein Generative Adversarial Networks (GAN) and Diffusion (DIFF) models, which are trained on datasets generated using Probabilistic Road Map (PRM) ~\cite{kavraki1996prm} planners in MATLAB. The code used for all the experiments can be found \href{https://github.com/franzjack/Certified_Generation_4_Planning.git}{here}. 

\subsection{Case Studies}
\begin{itemize}

    \item[(i)] \textbf{U-shaped Maze} -- \texttt{UMaze}: an agent navigating from left to right in a U-shaped maze. The objective is to move across the maze without colliding with the walls. 
    The model horizon is set to \(\bar{\horizon} = 24 \) for GAN and 17 for DIFF, with 
    \(h = 1\). 
    The STL specification encodes the safety constraints, i.e. always keep a safety distance from the walls: $        \property:= \square_{[0,\bar{\horizon}]} \big[
        \big(\|\vec{\mathbf{s}}-(25,15)\|_\infty \ge (5,15)\big) \land \big((5,5) \le \vec{\mathbf{s}} \le (45,45)\big)
        \big],$
    where $\vec{\mathbf{s}}$ denotes a trajectory.

    \item[(ii)] \textbf{Urban Road Intersection} -- \texttt{Crossroad}: an ego vehicle navigates an urban intersection with the option to turn left, proceed straight, or turn right, all while remaining within its designated lane. 
    The model horizon is set to \(\bar{\horizon} = 23\) for GAN and 16 for DIFF, with 
    \(h = 1\). 
   We choose a reach-avoid STL specification that encodes the safety constraints, i.e. always stay on the road and impose the generation of left turns:
    \begin{align*}
        \property:= \square_{[0,\bar{\horizon}]} 
        &\big[\wedge_{i=1}^4 
        \|\vec{\mathbf{s}}-c_i\|_\infty \ge \ell_i\big]\\ \nonumber
        &\land \Diamond_{[0,\bar{\horizon}]} \square_{[0,\bar{\horizon}]} \big[(0,27) \le \vec{\mathbf{s}} \le (15,35)\big],
    \end{align*}
   where $(c_i, \ell_i)$ denotes the centre and the radius of road corners.

\item[(iii)] \textbf{2D Navigation} -- \texttt{Obstacles}: an agent navigates from a fixed starting point to a predetermined destination point within a 2D space, avoiding any obstacles present. The task showcases multimodal behaviour, requiring the generative model to recognize and evaluate multiple potential paths that avoid obstacles. The model horizon is set to \(\bar{\horizon} = 15\), with 
\(h = 1\).
Once again, we choose a reach-avoid STL specification that lets the agent reach the goal while avoiding all the obstacles:
    \begin{align*}
        \property:= \square_{[0,\bar{\horizon}]} 
        &\big[\wedge_{i=1}^4 
        \|\vec{\mathbf{s}}-c_i\|_\infty \ge \ell_i\big]\\ \nonumber
        &\land \Diamond_{[0,\bar{\horizon}]} \square_{[0,\bar{\horizon}]} \big[\|\vec{\mathbf{s}}-(29,29)\|_\infty \le 1\big],
    \end{align*}
   where $(c_i, \ell_i)$ denotes the centre and the radius of the obstacles.

\item[(iv)] \textbf{Drone Navigation Scenario} -- \texttt{City}: a drone is required to fly over an urban landscape without ascending too high above the ground or crashing into buildings. 
The model horizon is set to \(\bar{\horizon} = 30\), with 
\(h = 1\). 
 The STL specification encodes the safety constraints, i.e. always keep a safety distance from the buildings: $
        \property:= \square_{[0,\bar{\horizon}]} 
        \big[\wedge_{i=1}^5
        \|\vec{\mathbf{s}}-c_i\|_\infty \ge \ell_i\big],$
     where $c_i$ denotes the centre of each building (or group of buildings and $\ell_i = (\ell_{i,x},\ell_{i,y},\ell_{i,z})$ the measures of each building.


\end{itemize}
\cref{fig:gan_planners} shows a visualization of the considered planning problems.

\subsection{Experimental Details}

The first experimental step lies in generating a pool of planning realization using PRM and using these as a dataset to train our DGM (both a GAN and a DIFF). Upon successful training of the DGMs, we proceed to test our certified guidance method. The initial step involves defining a Signal Temporal Logic (STL) property $\property$ that articulates the temporal behaviours we require our planner to enforce. This ensures that the outputs generated by the certified DGM, i.e. the DGM with a mixture of truncated Gaussians as latent distribution, will satisfy these requirements by design.

\subsubsection{Implementation Details}
\label{subsubsec:Implementation}
For NN verification, we leverage the flexibility and scalability of linear relaxation verification algorithms. In particular, we use Auto-LiRPA~\cite{xu2020automatic} (Automatic Linear Relaxation-based Perturbation Analysis), a tool designed to provide provable guarantees on neural network behaviours under input perturbations. Auto-LiRPA applies a linear relaxation technique to automatically compute bounds for the outputs of neural networks in response to perturbations in the input space. 
In our methodology, we employ AutoLiRPA to assess how the satisfaction of STL properties varies inside a given hyper-rectangle in the latent space. In particular, we utilize the CROWN (Convex Relaxation Over Weighted Networks) strategy as Interval Bound Propagation (IBP) proved to be ineffective, yielding bounds that were not useful in practice due to their conservative nature.
A considerable effort lies in encoding GAN and DIFF models together with the Boolean STL semantics into an AutoLiRPA-compliant format. The Boolean STL semantics is integrated into the computational graph of the DGM, allowing for perturbations in the latent space to propagate through to the satisfaction space, ensuring that the generated trajectories adhere strictly to the defined STL requirements.
For STL guidance, we implemented a differentiable quantitative semantics for STL within the PyTorch framework. This approach enables the application of gradient ascent techniques to identify latent inputs that maximize the satisfaction of the STL properties.

\begin{table}
\centering
\caption{Comparison of Acceptance Ratio: number of samples needed to obtain a satisfying trajectory across different methods and scenarios.}
\label{tab:computational_times}
\resizebox{\columnwidth}{!}{ 

\begin{tabular}{lp{1.5cm}p{1.5cm}p{1.5cm}p{1.5cm}p{1.5cm}}
\toprule
\textbf{GAN}  & \textbf{UMaze} & \textbf{Crossroad} & \textbf{Obstacles} & \textbf{City} \\
\midrule
Original      & 0.08           & 0.27               & 0.11               & 0.47 \\
Guidance      & 1.00           & 0.72               & 0.96               & 0.99 \\
Certified    & 1.00           & 1.00               & 1.00               & 1.00 \\
\bottomrule
\toprule
\textbf{DIFF} & \textbf{UMaze} & \textbf{Crossroad} & \textbf{Obstacles} & \textbf{City} \\
\midrule
Original      & 0.26    & 0.36        &        0.09         & - \\
Guidance      & 0.87    & 1.00       &       0.56           & - \\
Certified    & 1.00           & 1.00               & 1.00               & - \\
\bottomrule
\end{tabular}
}
\end{table}




\begin{table}
\centering
\caption{Comparison of Log Likelihoods: sum of the log-likelihoods w.r.t. $p(\latentvar)$ over $200$ samples across different methods and scenarios.}
\label{tab:log_likelihoods}
\resizebox{\columnwidth}{!}{ 
\begin{tabular}{lp{1.5cm}p{1.5cm}p{1.5cm}p{1.5cm}p{1.5cm}}
\toprule
\textbf{GAN}  & \textbf{UMaze} & \textbf{Crossroad} & \textbf{Obstacles} & \textbf{City} \\
\midrule
Original      &  -6452.1       & -8783.7    & -3981.3            & -8530.5 \\
Original Sat. &  -6729.3       & -8902.1    & -4049.3            & -8494.7 \\
Guidance      & -8955.7        & -9436.7    & -4244.8            & -14152.7 \\
Certified    & -7062.9        & -8295.7    & -3740.2            & -9469.2 \\
\bottomrule
\toprule
\textbf{DIFF} &\textbf{UMaze} & \textbf{Crossroad} & \textbf{Obstacles} & \textbf{City} \\
\midrule
Original      &       -8823.4        &          -8524.9          &    -7944.6                & - \\
Original Sat.    &      -8853.7         &       -8452.9             &    -7885.4                & - \\
Guidance      &        -8902.3       &     -8459.4              &           -7685.2         & - \\
Certified    &        -9810.5       &        -8691.9            &           -8515.6         & - \\

\bottomrule
\end{tabular}
}
\end{table}



\paragraph{DGM Architectural detail.}  For GANs, we use the same set of hyperparameters for all the case studies except for the dimension of the latent input that we choose to match the length of the considered trajectories. 
Similarly, we keep all the hyperparameters fixed also for the DIFF experiments. In the diffusion process, we set the minimum noise level $\beta_1=0.0001$ and the maximum noise level $\beta_{T}=0.5$, while the other noise levels are set using the following quadratic policy,
    $\beta_{\tau} = \left(\beta_{\tau}(T -\tau)/(T-1) - \beta_1(\tau-1)/(T - 1)    \right)^2$,
    as it has been shown that this type of decay 
    improves sample quality \cite{nichol2021improveddiff, song2020score, tashiro2021csdi}. As it greatly affects the computational burden of the verification process, we choose a low number $T$ of diffusion steps, in particular, we set $T=6$ for all the experiments. 
    After training the DDPM, we generate samples from the related DDIM by setting $\sigma_{\tau} = 0$ for all $\tau$ in the backward process; i.e., we remove the stochasticity in the generation process, thus making verification feasible.
In all the experiments, the Rectified ADAM (RADAM) algorithm \cite{liu2021radam} with a learning rate of $0.0005$ has been used to optimize the loss function. Training times depend heavily on the number of iterations, the number of training instances and their dimensionality.

\begin{figure}
    \centering
    \includegraphics[width=\columnwidth]{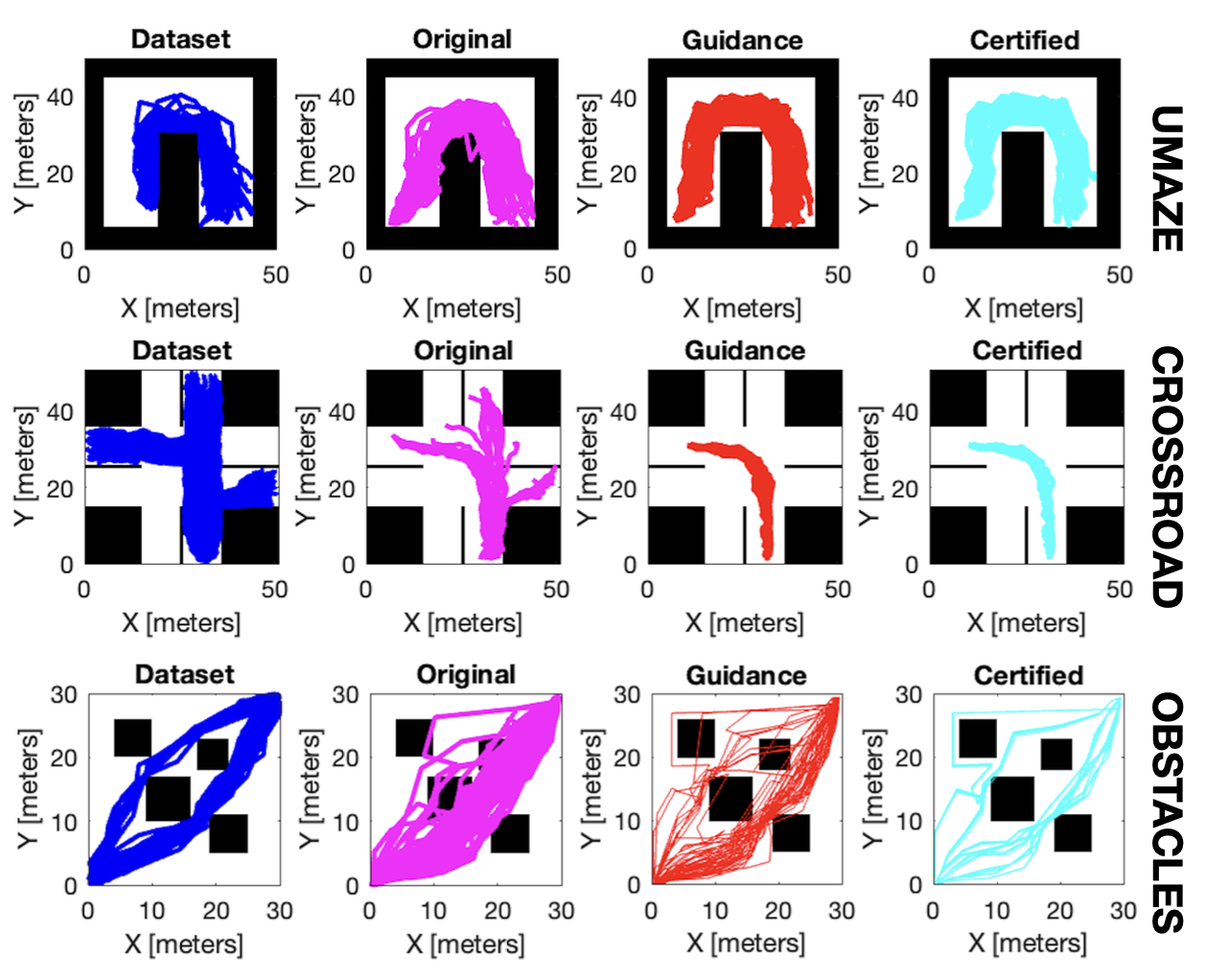}
   \includegraphics[width=0.8\columnwidth]{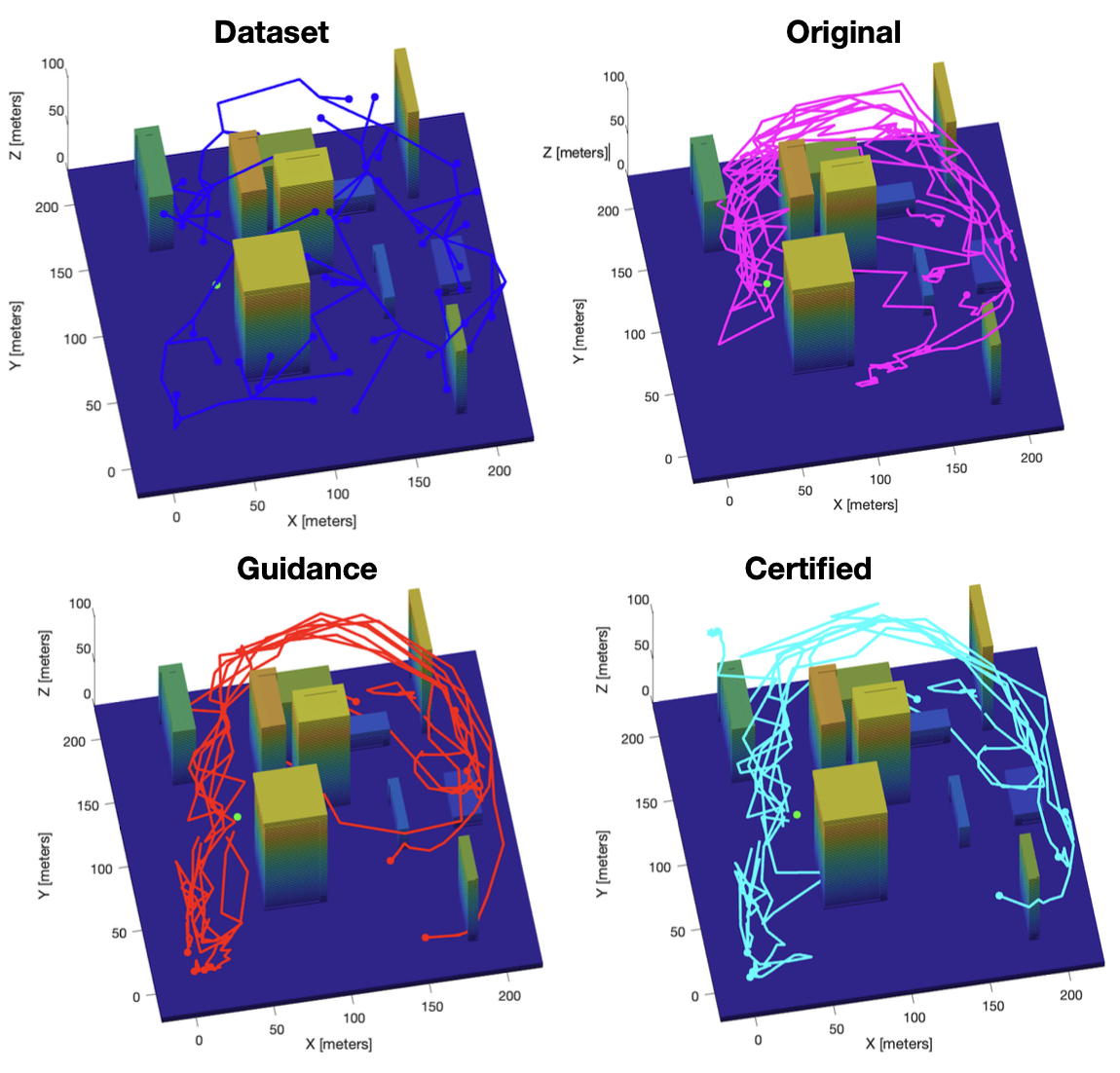}
    \caption{Visualizations of GAN-generated paths using each approach on different benchmarks.\vspace{-.25cm}}
    \label{fig:gan_planners}
\end{figure}

\paragraph{Experimental settings.} 
Each iteration of our experimental procedure begins with an initial perturbation where \( \epsilon = 0.01 \) and an incremental increase \( \Delta\epsilon = 0.005 \). 
The iterative process is repeated for a total of \( M = 20 \) cycles for GAN and \( M = 5 \) for DIFF, implying that at most 20 and 5 hyper-rectangles are generated respectively throughout the experiment. This ensures a structured approach to exploring the robustness and variability of our model outputs under controlled perturbations.
We experimented with heterogeneous increments of the hyper-rectangles but it was less effective at identifying satisfying regions. So, we report results only for the homogeneous increment version. We suspect that the gradient may cause abrupt increases in a latent space that is particularly stiff.


\subsection{Results}

\paragraph{Computational costs.}
\cref{tab:computational_times} computes, for each method, the ratio between 200 and the number of samples needed to obtain 200 satisfying trajectories. Samples from the original DGM show a low acceptance ratio, meaning that such samples have a low probability of satisfying $\property$. This happens also in scenarios where the DGM was trained over data that contained no violations, meaning that such violations were introduced by the generative model. This can be seen in the Obstacle Case Study (see \cref{fig:gan_planners}), where all training trajectories (in blue) satisfy the planning requirements, whereas samples from the original DGM (in purple) often violate the property. This comes with no surprises as the DGM generalizes over the possible dynamics with no explicit information about the regions to avoid. STL guidance shows a much higher acceptance ratio over all the case studies; however, it does not guarantee a 100\% satisfaction rate, which is instead achieved by construction by our certified DGM. 
From a computational point of view, the runtime needed to generate 200 $\property$-satisfying samples from the original models takes on average 3 seconds (15 seconds for DIFF), around half a second (2 seconds for DIFF) for our certified DGM  and around 20 minutes (60 minutes for DIFF) for the guidance approach where we have to perform 200 gradient searches, i.e. solve 200 optimization problems. Add times for DIFF.
Some offline costs need to be taken into account. The training time for the generative models is a cost shared by all the methods. However, our certification approach has a considerable computational overhead due to the search for good pivot points and the NN verification step. The cost of searching for the pivot is negligible; it takes less than 10 seconds. Performing one iteration of NN verification, however, takes $\sim 2$ minutes for GANs and $\sim 120 $ minutes for DIFFs. 
This significative difference in verification time depends on the dimensionality of the latent spaces and on the different architectural complexity of the two models. 
Diffusion models have a convoluted structure and the latent space is forced to have the same dimensionality of the output, whereas GAN does not have such a constraint.
These offline costs accumulate as we perform an iterative expansion of our hyper-rectangles. However, \cref{sec:algorithm}'s algorithm can be parallelized so that the $M$ hyper-rectangles can be computed in parallel, checking for pair-wise intersections only at the end.
The verification time for DIFF grows exponentially with the complexity of the model and the dimensionality of the associated latent space. DIFF faces a clear scalability issue. We were not able to perform experiments on the three-dimensional case study (\texttt{City}) due to the larger latent space (size $3\cdot \horizon$) and the increased number of diffusion steps needed to achieve a well-behaved generative model. 

\begin{figure}
    \centering
    \includegraphics[width=\columnwidth]{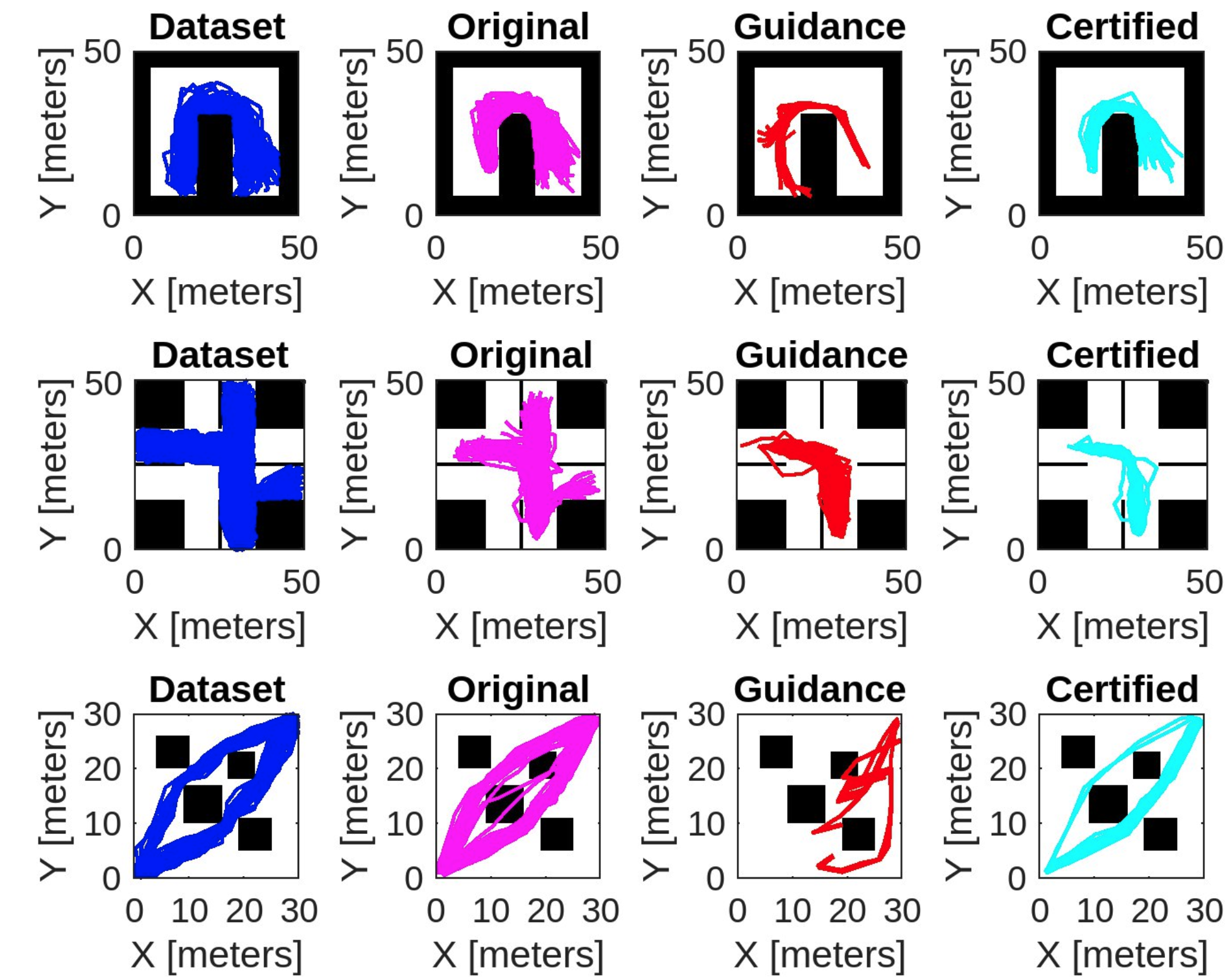}
    \caption{DIFF Certified paths in different benchmarks\vspace{-.25cm}}
    \label{fig:diff_planners}
\end{figure}

\paragraph{Goodness of fit.} \cref{tab:log_likelihoods} shows a comparison of the cumulative log-likelihood w.r.t. original latent distribution \( p(\latentvar\mid\condvar) \) of samples coming from the STL guidance and samples drawn from the new certified distribution \( p_{\property}(\latentvar\mid\condvar) \). As a baseline, we show the log-likelihood of samples taken from the original latent distribution and compare it with the likelihood of those samples that satisfy $\property$. More specifically, we will compare the sum of the log-likelihoods over 200 samples. Despite our certified guidance method considers only a small subset of the support, it results in trajectories that are as likely or more likely than the simple guidance method. 

\paragraph{Performances.}
For each case study, \cref{fig:gan_planners} for GAN and \cref{fig:diff_planners} for DIFF present a qualitative comparison of trajectories sampled either from the test dataset (in blue), from the original DGM (in purple), via STL guidance sampling (in red), and trajectories produced by our certified DGM (in cyan). 
These figures serve as a reference point for a qualitative evaluation of the performances of our certified guidance. In general, we notice how the original DGM introduces violations even when trained on a dataset that contained none. For instance, the \texttt{UMaze} and the \texttt{Obstacles} case studies show how none of the training trajectories (blue) crashed into an obstacle, whereas the DGM (purple) does. STL guidance stirs the sampling towards regions of satisfaction but does not guarantee that it produces violation-free trajectories. For instance, in the GAN \texttt{Crossroad} scenario, we had to reject $28\%$ of the samples obtained via STL guidance because they did not satisfy the STL requirement. 

\paragraph{GAN Performance.} Our GAN-certified guidance approach offers good overall results in terms of variability of the generated trajectories. Its strengths lie in the sampling efficiency at runtime and in the insights we gain over how the latent space encodes information. For instance, in the \texttt{Crossroad} scenario, one could distinguish the areas of the latent space that generate left turns from those that generate right turns. Moreover, we expect these regions to correspond to robust encodings of the dynamics as contiguous latent points lead to similar outputs. Moreover, we preserve the original likelihood ratio so that latent inputs with low probability are not sampled too often. In general following the gradient of the robustness, as STL guidance does, may lead far from high probability regions. 
However, we can guarantee that we are not over-sampling from these regions, whereas STL guidance cannot offer such guarantees. Moreover, Certified GAN also performs well in the three-dimensional planning problem (\texttt{City})
, showing a good level of scalability.

\paragraph{DIFF Performance.} 
Due to the aforementioned scalability issues of NN verification, we must keep the latent space to a manageable dimension. Therefore, DIFF models are trained on trajectories shorter than the ones used by GANs. For all the considered Case Studies, we downsample the original PRM trajectories. 
The number of diffusion steps was deliberately kept low to ensure feasible verification times. Moreover, we are sampling from the implicit version of a diffusion model, which introduces stochasticity only in the initial step, drastically reducing the variance of the generated samples. We also had to impose shallower STL constraints to allow the STL guidance to converge to positive robustness values. Although this made verification possible, it still took considerably more time than GAN-based verification, and exploring the latent space remains challenging. This is further supported by the low acceptance ratios of the STL guidance. The gradient search tends to fall into the same regions consistently; and almost certainly due to the mentioned constraints, the latent space does not encode information robustly. Thus, finding acceptable hyper-rectangles is more difficult. While the results could be stronger, this is a first step in this direction, and our approach provided valuable insights into the behavior of the diffusion model w.r.t. its latent space.


\vspace{-0.2em}
\section{CONCLUSION}\label{sec:conclusion}

We introduced certified guidance, a novel framework to ensure that the outputs of deep generative models satisfy specific planning constraints. Our method is inspired by gradient-based approaches to guidance but, unlike these, can provide rigorous correctness guarantees. 
Our experiments have shown promising results, demonstrating the reliability of our certified models in cases where ``vanilla'' guidance fails. We managed to certify both GAN and diffusion models, albeit the latter suffer from scalability issues. Despite that, to our knowledge, we are the first to successfully certify diffusion models w.r.t.\ temporal logic requirements. 
Additionally, this framework can be directly extended to VAEs, where a variational decoder can adopt the same structure as the GAN generator, further broadening its applicability. 
In future work, we aim to reduce the dependence on specific values in the conditioning space. We will focus on developing guarantees that generalize across the entire conditioning space, allowing for more flexible and reliable planning outputs. This direction will further enhance the robustness and practical usability of certified guidance in real-world scenarios.




\begin{acks}
This work is supported by the European Union's HORIZON-CL4-HUMAN-01 programme under grant agreement 101070028,  the Engineering and Physical Sciences Research Council under Award EP/W014785/2 and by the PNRR project iNEST (Interconnected North-Est Innovation Ecosystem) funded by the European Union Next-GenerationEU (Piano Nazionale di Ripresa e Resilienza (PNRR) – Missione 4 Componente 2, Investimento 1.5 – D.D. 1058 23/06/2022, ECS\_00000043).
\end{acks}



\bibliographystyle{ACM-Reference-Format} 
\bibliography{references}

\clearpage

\begin{appendices}

\section{Background}

\subsection{Deep Generative Models}\label{app:dgm}

Every dataset can be considered a set of observations $\targetvar$ drawn from an unknown distribution $p(\targetvar)$. Generative models aim to learn a model that mimics this unknown distribution as closely as possible, i.e., learn a parametric distribution $p_\weights(\targetvar)$ that approximates $p(\targetvar)$ and allows for efficient sampling. 

A generative model acts as a distribution transformer, i.e., a map $\gen:\latentspace\to \targetspace$ transforming a simple distribution $p(\latentvar)$ over a latent space $\latentspace$ into a complex distribution $p_\weights(\targetvar)\approx p(\targetvar)$ over the target space $\targetspace$. In other words, they generate data from noise. 
In our planning applications, we want the generation of stochastic trajectories to be conditioned for instance on the starting position of the system or both the starting and terminal position of the system. To approximate such conditional distribution $p(\targetvar\mid \condvar)$, we instead use 
conditional generative models, which can be seen as maps $\gen: \latentspace\times \condspace\to \targetspace$, where $\condspace$ is the conditioning space. 
Training the generative model requires minimising a differentiable loss function $\loss$, 
which measures the difference between true and generated samples for each condition $\condvar$ in $\trainingset$. 

\subsubsection{Generative Adversarial Nets}\label{subsec:wgan}

Generative Adversarial Nets (GANs)~\cite{goodfellow2014generative} are a class of deep learning-based generative models. 
In this work we consider Wasserstein GANs (WGAN)~\cite{arjovsky2017wasserstein,gulrajani2017improved}, a popular variant of GANs that is more stable and less sensitive to the choice of model architecture and hyperparameters. 
WGANs use the Wasserstein distance 
to measure the difference between the learned distribution $p_\weights(\targetvar)$ and the target distribution $p(\targetvar)$. Because of  the Kantorovich-Rubinstein duality \cite{villani2008optimal} such distance can be computed as the supremum over all the 1-Lipschitz functions $f : \targetspace \rightarrow \mathbb{R}$:
\begin{equation}\label{eq:wassdist}
\small 
W(p,p_\weights) =  \sup_{||f||_L\le 1} \left(
\mathbb{E}_{\targetvar\sim p(\targetvar)}[f(\targetvar)]-\mathbb{E}_{\targetvar\sim p_\weights(\targetvar)}[f(\targetvar)]
\right).
\end{equation}
We approximate these functions $f$ with a neural net $C_{\weights_c}$, referred to as \textit{critic}, parametrized by weights $\weights_c$. 
To enforce the Lipschitz constraint we follow \cite{gulrajani2017improved} and introduce a penalty over the norm of the gradients. It is known that a differentiable function is 1-Lipchitz if and only if it has gradients with a norm at most 1 everywhere. 
The objective function, to be maximized w.r.t. $\weights_c$ and minimized w.r.t. $\weights$, becomes:
\begin{align}\label{eq:wassdist_gp}
\small 
\mathcal{L}_{\mathit{gan}}({\weights_c}, \weights) := \mathbb{E}_{\targetvar\sim p(\targetvar)}&[C_{\weights_c}(\targetvar)]-\mathbb{E}_{\targetvar\sim p_{\weights}(\targetvar)}[C_{\weights_c}(\targetvar)]\\&-\lambda \mathbb{E}_{\hat{\targetvar}\sim p_{\hat{\targetvar}}}
( \lVert \nabla_{\hat{\targetvar}}C_{\weights_c} (\hat{\targetvar})  \lVert_2-1 )^2]  ,\nonumber
\end{align}
where $\lambda$ is the penalty coefficient and $p_{\hat{\targetvar}}$ is defined by sampling uniformly along straight lines between pairs of points sampled from $p$ and $p_{\weights}$. This is actually a soft constraint, meaning it does not guarantee the Lipschitzianity of the critic, however, it performs well in practice~\cite{gulrajani2017improved}.
The synthetic data $\targetvar\sim p_{\weights}(\targetvar)$ is obtained via a \textit{generator} $G_{\weights} : \latentspace \rightarrow \targetspace$ which is trained to mimic the data distribution $p(\targetvar)$ by transforming a latent vector $\latentvar\sim p(\latentvar)$ (typically a standard Gaussian distribution).

When dealing with inputs that are trajectories, i.e. sequences of fixed length, convolutional neural networks (CNNs)~\cite{goodfellow2016deep} can be used for both the generator and the critic.

\paragraph{Conditional Generative Adversarial Networks} Conditional Generative Adversarial Nets (cGAN)~\cite{mirza2014conditional} are a type of GANs that involves the conditional generation of examples, i.e., the generator produces examples of a required type, e.g. examples that belong to a certain class, and thus they introduce control over the desired generated output. In our planning applications, we want the generation of stochastic trajectories to be conditioned for instance on the starting position of the system or both the starting and terminal position of the system.The architecture used in this work is thus a conditional Wasserstein Convolutional GAN with a gradient penalty.

\subsubsection{Denoising Diffusion Models}
Diffusion probabilistic models~\cite{sohldickstein2015} are the state-of-the-art approach to learn models of data generative distributions from data. In particular, we focus on the variant presented in~\cite{Ho2020DenoisingDP}, called denoising diffusion probabilistic models (DDPMs). DDPMs are equivalent, during the training phase, to score-based models at multiple noise levels with Langevin dynamics~\cite{song2020}.
Let $\targetvar^0$ denote samples coming from the unknown data distribution $\targetvar^0\sim p(\targetvar^0)$ over a space $\targetspace$ and let $\targetvar^\tau$ for $\tau = 1,\ldots,T$ be a sequence of variables in the same space $\targetspace$ of $\targetvar^0$. Let $p_\weights(\targetvar^0)$ be a distribution that approximates $p(\targetvar^0)$.
Diffusion probabilistic models are latent variable models composed of two processes: a forward and a reverse process. 

The \emph{forward process} is a Markov process designed to iteratively turn a sample of $p(\targetvar^0)$ into pure noise (typically a standard Gaussian distribution). It is  defined by the following Markov chain:
\begin{equation}\label{eq:forward}
p(\targetvar^{1:T}| \targetvar^0) = \prod_{\tau=1}^T p(\targetvar^\tau| \targetvar^{\tau-1}),
\end{equation}
where $p(\targetvar^\tau| \targetvar^{\tau-1}):=\mathcal{N}\left(\sqrt{1-\beta_\tau}\targetvar^{\tau-1},\beta_\tau \mathbb{I}\right)$ and $\beta_\tau$ is a positive constant representing the noise level introduced at diffusion step $\tau$. The conditional distribution $p(\targetvar^\tau| \targetvar^0)$ can be written in closed form as $p(\targetvar^\tau| \targetvar^0) = \mathcal{N}\left(\sqrt{\alpha_\tau}\targetvar^0,(1-\alpha_\tau) \mathbb{I}\right)$, where $\alpha_\tau = \prod_{i=0}^\tau (1-\beta_i)$. Thus, $\targetvar^\tau$ can be expressed as $\targetvar^\tau = \sqrt{\alpha_\tau}\targetvar^0+(1-\alpha_\tau)\epsilon$, where $\epsilon\sim\mathcal{N}(\mathbf{0},\mathbf{I})$.

The \emph{reverse process} inverts back in time the forward process, denoising $\targetvar^\tau$ to recover $\targetvar^0$.  It is defined by the following reverse Markov chain:
\begin{equation}\label{eq:reverse}
\begin{aligned}
&p_\weights(\targetvar^{0:T}) := p_\weights(\targetvar^{T})\prod_{\tau=1}^T p_\weights(\targetvar^{\tau-1}| \targetvar^\tau), \quad \targetvar^T\sim\mathcal{N}(\mathbf{0},\mathbf{I}) \\
&p_\weights(\targetvar^{\tau-1}| \targetvar^\tau) := \mathcal{N}\left(\targetvar^{\tau-1}; \mu_\weights(\targetvar^\tau,\tau),\sigma_\tau\mathbf{I}\right).
\end{aligned}
\end{equation}
In~\cite{Ho2020DenoisingDP} the following specific parameterization of $p_\weights(\targetvar^{\tau-1} |\targetvar^\tau)$ has been proposed:
\begin{equation}
\begin{aligned} 
    \mu_\weights(\targetvar^\tau,\tau) &= \frac{1}{\alpha_\tau}\left(\targetvar^\tau-\frac{\beta_\tau}{\sqrt{1-\alpha_\tau}}\epsilon_\weights(\targetvar^\tau,\tau)\right) \label{eq:ddpm_mu}\\
    \sigma^2_\tau &=     \begin{cases}
        \frac{1-\alpha_{\tau-1}}{1-\alpha_\tau}\beta_\tau,\quad \tau > 1\\
        \beta_1,\quad \tau = 1 
    \end{cases}
\end{aligned}
\end{equation}
where $\epsilon_\weights:\targetspace\times\mathbb{R}\to \targetspace$ is a trainable denoising function.
The reverse process is not analytically computable and it has to be trained from data by minimizing the following objective function w.r.t. $\weights$: 
\begin{equation}\label{eq:reverse_optim}
\mathcal{L}_{\mathit{diff}}(\weights) := \ \mathbb{E}_{\targetvar^0\sim p(\targetvar^0),\epsilon\sim\mathcal{N}(\mathbf{0},\mathbf{I}),\tau} ||\epsilon-\epsilon_\weights(\targetvar^\tau,\tau) ||^2_2.
\end{equation}
The denoising function $\epsilon_\weights$ estimates the noise vector $\epsilon$ that was added to its noisy input $\targetvar^\tau$. 
This training objective can also be viewed as a weighted combination of denoising score matching used for training score-based generative models~\cite{song2019generative,song2020score,song2020improved}. Once trained, we can sample $\targetvar^0$ from~\eqref{eq:reverse}.
So the sample generating function $G_\weights$ can be defined as the concatenation of sampling procedures along the $T$ diffusion steps.
The sampling can be formalized as $\bar{\targetvar}^T\sim \mathcal{N}(\mathbf{0},\mathbf{I})$, $\bar{\targetvar}^{T-1}\sim p_\weights(\targetvar^{T-1}| \bar{\targetvar}^T)$ until $\bar{\targetvar}^0\sim p_\weights(\targetvar^0| \bar{\targetvar}^1)$ which is our desired sample in $\targetspace$. In this formalization, $\targetvar^0$ denotes our target data $\targetvar$ with unknown distribution $p(\targetvar)$, whereas $\targetvar^T$ corresponds to our latent input $\latentvar$ with latent distribution $p(\latentvar)=\mathcal{N}(\mathbf{0},\mathbf{I})$.

\paragraph{Conditional Denoising Diffusion.}
Given a sample $\targetvar^0$ whose observation is conditional on a variable $\condvar$, we need to consider a conditional diffusion model that estimates the probability $p(\targetvar^0|\condvar)$.
The reverse process aims at modeling $p(\targetvar^{\tau-1}|\targetvar^\tau,\condvar)$. The denoising function becomes $\epsilon_\weights:\left(\targetspace\times\mathbb{R}\mid \condspace\right)\to \targetspace$ and the optimization becomes

\begin{equation}
\label{eq:cond_reverse_optim}
\min_\weights\ \mathbb{E}_{(\targetvar^0,\condvar^0)\sim p(\targetvar^0,\condvar),\epsilon\sim\mathcal{N}(\mathbf{0},\mathbf{I}),\tau} ||\epsilon-\epsilon_\weights(\targetvar^\tau,\tau|\mathbf{y} ||^2_2.
\end{equation}

\paragraph{Implicit Denoising Diffusion Models.}
The computational burden and complexity of sampling from a trained DDPM can be reduced by resorting to their implicit formulation~\cite{song2022implicit}.
The so-called Denoising Diffusion Implicit Models (DDIM) were first introduced to speed up the sample-generating procedure, however, they present the additional advantage of removing all the stochasticity in the intermediate steps of the denoising process. Consider the standard probabilistic formulation provided in Eq.~\eqref{eq:forward} and ~\eqref{eq:reverse} and let $\tilde{\epsilon}_{\weights}(\targetvar^{\tau}, \tau)  =  \frac{ \sqrt{1- \alpha_{\tau}}\cdot\epsilon_{\weights}(\targetvar^{\tau}, \tau)}{\sqrt{\alpha_{\tau}}}$. Within this formulation, in the generative process sample $\targetvar_{\tau - 1}$ can be obtained from $\targetvar_{\tau}$ via:

\begin{equation}
\label{eq:rev_diff}
    \targetvar_{\tau-1} =  \Bigg(\frac{\targetvar^{\tau} }{\sqrt{1- \beta_{\tau}}}\Bigg) - \tilde{\epsilon}_{\weights}(\targetvar^{\tau}, \tau)  + \Big(\sqrt{1 - \alpha_{\tau-1} - \sigma_{\tau}^2}\Big) \cdot \epsilon_{\weights}(\targetvar^{\tau}, \tau) + \sigma_{\tau}\epsilon_\tau,
\end{equation}
where $\epsilon_\tau\sim\mathcal{N}(\mathbf{0},\mathbf{I})$.

When $\sigma_\tau=\sqrt{(1-\alpha_{\tau-1})/(1-\alpha_\tau)}\sqrt{1-\alpha_\tau/\alpha_{\tau-1}}
$ for all $\tau$,
the generative process becomes a DDPM. On the other hand, if $\sigma_\tau = 0$ for all $\tau$, the generative process becomes a DDIM and the sampling procedure can be rewritten as:
\begin{equation}
\label{eq:rev_implicit}
    \targetvar_{\tau-1} =  \Bigg(\frac{\targetvar^{\tau} }{\sqrt{1-\beta_{\tau}}}\Bigg) - \tilde{\epsilon}_{\weights}(\targetvar^{\tau}, \tau) + \Big(\sqrt{1 - \alpha_{\tau-1}}\Big) \cdot \epsilon_{\weights}(\targetvar^{\tau}, \tau),
\end{equation}
thus making the backward process deterministic from the noisy latent to the denoised data-like samples $\mathbf{x}^0$. The main result presented in~\cite{song2022implicit} states that one can train a DDPM and then generate samples from the equivalent DDIM,  because the two models share the same loss.


\end{appendices}

\end{document}